\pdfoutput=1

\documentclass[11pt]{article}

\usepackage[preprint]{acl}

\usepackage{times}
\usepackage{latexsym}

\usepackage[T1]{fontenc}

\usepackage[utf8]{inputenc}

\usepackage{microtype}

\usepackage{inconsolata}

\usepackage{graphicx}
\usepackage{amsmath}
\usepackage{amssymb}
\usepackage{dsfont}
\usepackage{multirow}
\usepackage{multicol}
\usepackage{amscd}
\usepackage{tikz-cd}
\usepackage{booktabs}
\usepackage{subfigure}
\usepackage{float}
\usepackage{CJKutf8}

\title{Instructions for *ACL Proceedings}
\title{MEMLA: Enhancing Multilingual Knowledge Editing with Neuron-Masked Low-Rank Adaptation}

\author{
 \textbf{Jiakuan Xie\textsuperscript{\rm 1,\rm2}},
 \textbf{Pengfei Cao\textsuperscript{\rm 1,\rm2}},
 \textbf{Yuheng Chen\textsuperscript{\rm1,\rm2}},
 \textbf{Yubo Chen\textsuperscript{\rm 1,\rm 2}},
 \textbf{Kang Liu\textsuperscript{\rm 1,\rm 2}},
 \textbf{Jun Zhao\textsuperscript{\rm 1,\rm 2}}
\\
 \textsuperscript{1}The Laboratory of Cognition and Decision Intelligence for Complex Systems, \\
 Institute of Automation, Chinese Academy of Sciences, Beijing, China \\
 \textsuperscript{2}School of Artificial Intelligence, University of Chinese Academy of Sciences, Beijing, China
\\
\{xiejiakuan2023, chenyuheng22\}@ia.ac.cn, \{pengfei.cao, yubo.chen, kliu, jzhao\}@nlpr.ia.ac.cn 
}

\begin{document}
\maketitle
\begin{abstract}
Knowledge editing aims to adjust the knowledge within large language models (LLMs) to prevent their responses from becoming obsolete or inaccurate.
However, existing works on knowledge editing are primarily conducted in a single language, which is inadequate for multilingual language models. 
In this paper, we focus on multilingual knowledge editing (MKE), which requires propagating updates across multiple languages. This necessity poses a significant challenge for the task.
Furthermore, the limited availability of a comprehensive dataset for MKE exacerbates this challenge, hindering progress in this area.
Hence, we introduce the \textbf{M}ultilingual \textbf{K}nowledge \textbf{E}diting \textbf{B}enchmark (MKEB), a novel dataset comprising 12 languages and providing a complete evaluation framework. 
Additionally, we propose a method that enhances \textbf{M}ultilingual knowledge \textbf{E}diting with neuron-\textbf{M}asked \textbf{L}ow-Rank \textbf{A}daptation (MEMLA).
Specifically, we identify two categories of knowledge neurons to improve editing precision. Moreover, we perform LoRA-based editing with neuron masks to efficiently modify parameters and facilitate the propagation of updates across multiple languages.
Experiments demonstrate that our method outperforms existing baselines and significantly enhances the multi-hop reasoning capability of the edited model, with minimal impact on its downstream task performance. 
The dataset and code will be made publicly available.
\end{abstract}

\section{Introduction}

Transformer-based (\citealp{attentionisall}) large language models (LLMs) are capable of implicitly internalizing a wide range of knowledge during pretraining (\citealp{alkhamissi2022review}; \citealp{petroni-etal-2019-language}).
However, the potential for generating inaccurate and outdated responses limits the widespread applications of LLMs. 
One proposed solution to this problem is knowledge editing, which modifies specific factual knowledge in LLMs while ensuring no additional impact on other unrelated facts. This task allows for efficient alterations to language models without full retraining, thereby reducing computational costs.
Despite notable successes in knowledge editing across various studies (\citealp{editing-factual-in-LM}; \citealp{knowledge-neurons}; \citealp{rome}; \citealp{memit}; \citealp{mend}; \citealp{SERAC}; \citealp{wang2024easyedit}), the research has been conducted in a single language, where the source and target languages are identical. 
\begin{figure}[t]
    \centering
    \includegraphics[width=\columnwidth]{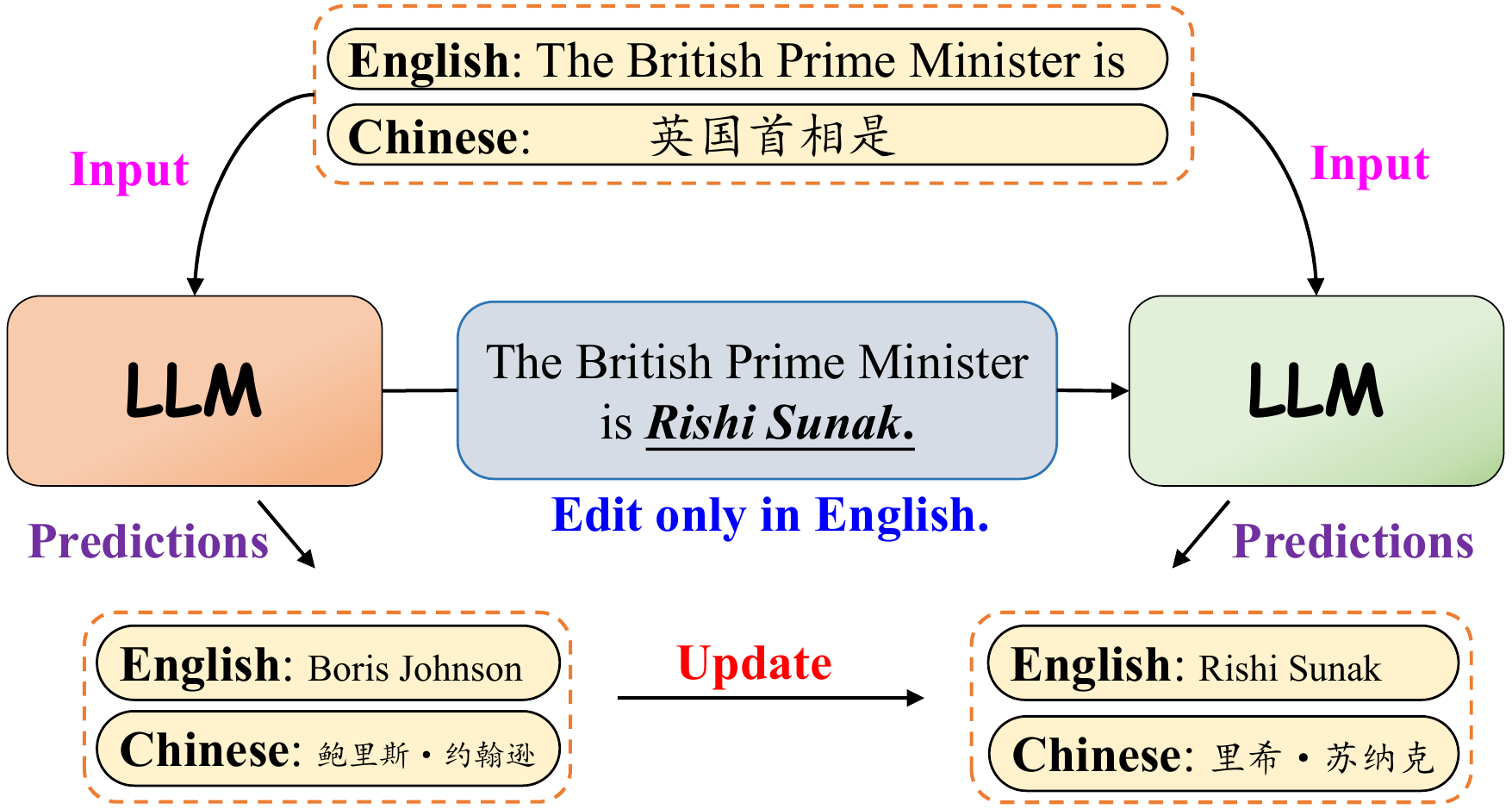}
    \caption{An example of MKE: when a fact is updated in one language (e.g., English), the new fact is transferred to other languages (e.g., Chinese).}
    \label{fig:introduction}
\end{figure}
As LLMs are required to handle and respond to queries in multiple languages on various tasks (\citealp{lm-multilingual-cot}; \citealp{mllm4mspeech}), it is imperative to advance knowledge editing from monolingual to multilingual settings to ensure that edited models can generate accurate responses to queries in various languages.
\textbf{M}ultilingual \textbf{K}nowledge \textbf{E}diting (MKE) requires modifications to be made in one language, accompanied by corresponding adjustments in multiple other languages. 
As shown in Figure~\ref{fig:introduction}, the anticipated outcome is that altering the British Prime Minister from ``Boris Johnson'' to ``Rishi Sunak'' in English would lead the language model to predict ``\begin{CJK}{UTF8}{gbsn}\small里希·苏纳克\end{CJK}'' (Rishi Sunak) in Chinese when presented with the corresponding Chinese query.
This requirement of transferring new knowledge across diverse languages poses a significant challenge to MKE, which has not been effectively addressed in existing works (\citealp{anisotropic}; \citealp{wang2023crosslingual}; \citealp{wang2023retrievalaugmented}; \citealp{mlake}).
Furthermore, there is currently a lack of a dataset to evaluate the reliability, generality, locality, cross-lingual transferability of editing algorithms, and the multi-hop reasoning capability of edited models\footnote{We compare several existing datasets in Table~\ref{tab:dataset-comparison}.}.

To address the aforementioned issues, we propose a novel benchmark, MKEB, encompassing 12 distinct languages. 
Each instance within the dataset for each language includes an edit prompt, paraphrase prompts, and neighborhood prompts. 
They are used for reliability evaluation, generality evaluation, and locality evaluation, respectively. 
Based on them, we can assess cross-lingual transferability.
Additionally, MKEB provides multi-hop questions to evaluate the multi-hop reasoning capability of edited models. 
Thus, our dataset allows for a comprehensive assessment.
Experiments conducted on this dataset demonstrate that an existing popular method, MEMIT (\citealp{memit}), achieves 99.45\% of reliability in monolingual setting, while only achieving an average of 58.46\% in cross-lingual settings, revealing that current methods face significant challenges in MKE scenarios.

In this paper, we propose a method that enhances multilingual knowledge editing with neuron-masked Low-Rank Adaptation (MEMLA). 
To improve editing precision, we identify two categories of knowledge neurons: language-specific knowledge neurons associated with a particular language and language-independent knowledge neurons that transmit knowledge in a more universal manner. 
To efficiently update parameters and facilitate the propagation of updates across multiple languages, we create neuron masks for Low-Rank Adaptation (LoRA) (\citealp{loramodule}) to adjust only the parameters associated with knowledge neurons in the Multi-Layer Perceptrons (MLPs). 
Thus, we achieve more precise, flexible, and lightweight modifications.
Experiments on our benchmark indicate that MEMLA exhibits superior performance compared to other baselines. 
Moreover, our method enhances the model's ability to effectively integrate new knowledge for multi-hop reasoning while causing minimal disruption.

Overall, the contributions of this paper can be summarized as follows:

\begin{itemize}
    \item We introduce a novel benchmark, MKEB, specifically designed for the Multilingual Knowledge Editing (MKE) task. This dataset encompasses 12 different languages and provides a comprehensive evaluation framework for reliability, generality, locality, transferability of editing algorithms, and multi-hop reasoning capability of edited models.
    \item We propose an effective multilingual knowledge editing approach called MEMLA. To improve editing precision, we identify two types of knowledge neurons. To efficiently update parameters and facilitate the propagation of updates into multiple languages, we create neuron masks for LoRA to adjust the critical parameters of a language model.
    \item We have conducted a series of experiments, and the results substantiate the superior performance of our approach compared to existing baselines. MEMLA achieves a 7.14\% improvement in average performance for cross-lingual settings and a 13.95\% improvement specifically when the source language is Chinese and the target language is Russian.
    Additionally, our method has proven effective in facilitating the edited model to perform multi-hop reasoning with minimal impact on its general performance. The dataset and code will be released publicly.
\end{itemize}

\section{Related Work}
\paragraph{Knowledge Editing.} The aim of knowledge editing is to modify the knowledge within LLMs to ensure their behaviors align with real-world facts. 
Currently, there are several paradigms for the task (\citealp{surveyofedit}):
(1) \textit{Memory-based Model}, which leaves the original model unchanged and influences the model output by retrieving related examples (\citealp{SERAC}; \citealp{memprompt}; \citealp{iclchangbaobao}; \citealp{MQuAKE}). 
(2) \textit{Additional Parameters}, which introduces extra learnable parameters within LLMs while preserving model parameters (\citealp{calinet}; \citealp{grace}; \citealp{TPatcher}).
(3) \textit{Locate-Then-Edit}, which identifies the related parameters within LLMs and adjusts them (\citealp{knowledge-neurons}; \citealp{rome}; \citealp{memit}). 
(4) \textit{Meta-learning}, which utilizes a hyper network to obtain parameter modifications (\citealp{mend}; \citealp{editing-factual-in-LM}).

\begin{figure*}[h]
    \centering
    \scalebox{0.85}{
        \includegraphics[width=\textwidth]{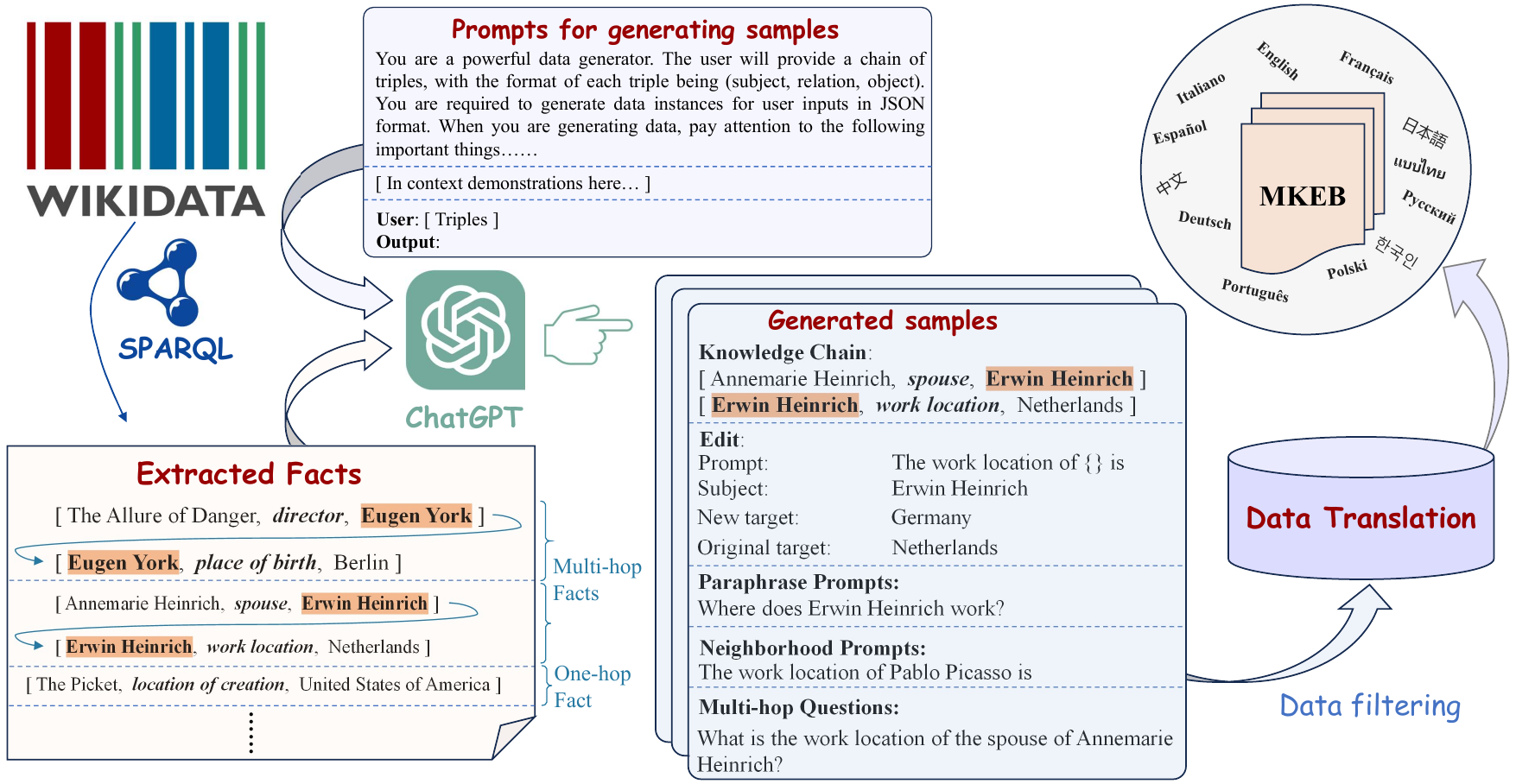}
    }
    \caption{The construction process of our MKEB dataset, which involves retrieving numerous facts from Wikidata, using handcrafted prompts to induce ChatGPT to generate data samples, and further processing these samples through filtering and translation.}
    \label{fig:dataset-construction}
\end{figure*}

\paragraph{Multilingual Knowledge Editing.} A distinctive feature of multilingual knowledge editing (MKE) is its requirement for alterations in one language to propagate to other languages. 
\citet{anisotropic} introduced language anisotropic editing, which amplifies different subsets of parameters for each language. 
\citet{wang2023crosslingual} constructed a dataset called Bi-ZsRE
and evaluated various knowledge editing methods based on it to study the cross-lingual effect.
\citet{beniwal2024crosslingual} highlighted the limitations of current knowledge editing techniques in MKE.
\citet{wang2023retrievalaugmented} introduced a retrieval-augmented multilingual knowledge editor that involves multilingual knowledge retrieval and multilingual in-context editing.
\citet{mlake} developed a multilingual dataset specifically for multi-hop reasoning and discovered that existing methods are limited in MKE.

Despite these successful efforts, several limitations remain.
Primarily, existing datasets support very few languages and provide a limited evaluation framework. 
Moreover, current methods struggle to effectively modify model parameters and transfer new knowledge across diverse languages.

\section{Dataset}
In this section, we provide a detailed explanation of the dataset construction (\S\ref{subsec:dataset-construction}) and the data statistics of our dataset (\S\ref{subsec:data-statistics}). The overall process of dataset construction is illustrated in Figure~\ref{fig:dataset-construction}.

\subsection{Dataset Construction}\label{subsec:dataset-construction}
\paragraph{Fact Extraction.} We represent a fact as a triple in the form of $\left(s, r, o\right)$ and utilize SPARQL to retrieve factual triples from Wikidata\footnote{\url{https://query.wikidata.org/}}. 
To create multi-hop questions that evaluate the multi-hop reasoning capability of edited models, we also retrieve knowledge chains consisting of multiple triples.

\paragraph{Sample Generation.} We utilize the ChatGPT API\footnote{\url{https://platform.openai.com/docs/guides/text-generation/chat-completions-api}} to generate samples. Specifically, we feed handcrafted prompts and demonstrations into ChatGPT, leveraging its in-context learning (\citealp{icl-survey}) capability to generate corresponding samples.
To ensure that the responses of ChatGPT align with our specified criteria, we meticulously craft generation guidelines provided in the input prompt. These guidelines are detailed in Appendix~\ref{sec:appendix-dataset}.

\paragraph{Data Filtering.} We conduct additional processing on the raw data generated by ChatGPT. Specifically, we select samples that fail to meet the specified criteria, provide comprehensive explanations for the necessary corrections, and regenerate them. Instances that have been generated more than three times and still fail to meet the specified requirements are excluded from the dataset.

\paragraph{Data Translation.} To acquire a multilingual dataset, we translate the processed data into multiple languages using the Baidu Translate API\footnote{\url{https://fanyi-api.baidu.com/}}, resulting in a final dataset available in 12 languages: English (en), Chinese (zh), French (fr), German (de), Japanese (ja), Korean (ko), Portuguese (pt), Russian (ru), Italian (it), Spanish (es), Polish (pl), and Thai (th).

\begin{figure}[t]
    \centering
    \includegraphics[width=\columnwidth]{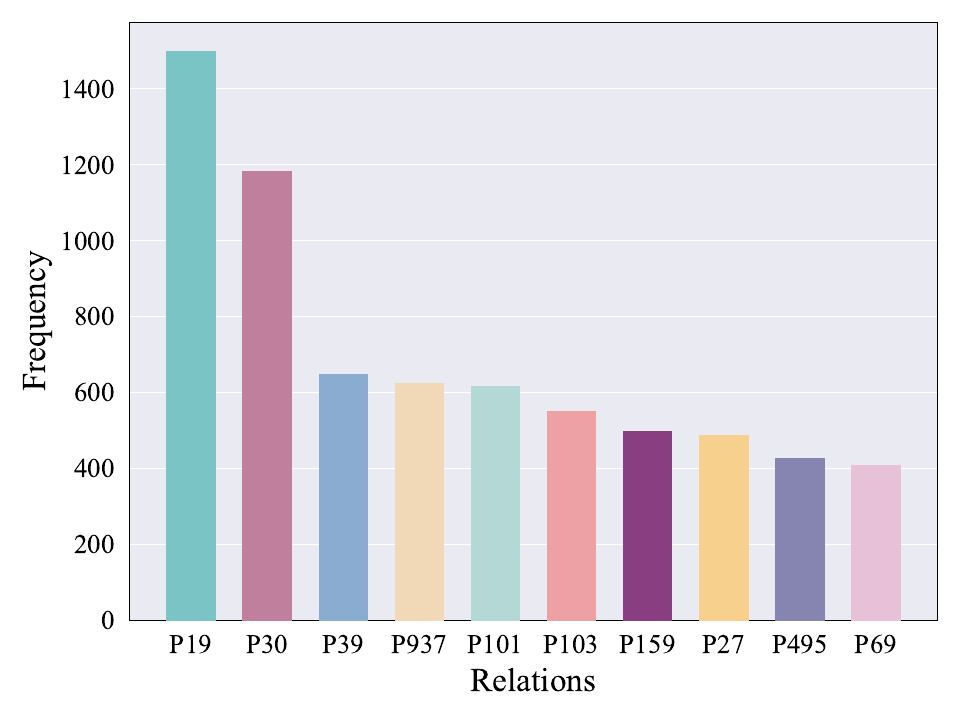}
    \caption{Distributions of top-10 relations in our dataset.}
    \label{fig:data-relation}
\end{figure}
\begin{table}[t]
    \centering
    \scalebox{0.75}{\begin{tabular}{c|cccc}
    \toprule
         Languages & Edit & Paraphrase & Neighborhood & MQ \\
         \midrule
         en & 9.19  & 10.69 & 8.16 & 15.67 \\
         zh & 9.35 & 12.58 & 8.74 & 16.13 \\
         fr & 10.03 & 12.55 & 9.16 & 16.66 \\
         de & 9.80 & 11.55 & 8.26 & 14.98 \\
         ja & 9.83 & 14.16 & 9.03 & 17.75 \\
         \bottomrule
    \end{tabular}}
    \caption{Average lengths of different types of prompts. Edit, Paraphrase, and Neighborhood represent three distinct types of prompts, while MQ indicates multi-hop questions. The languages listed in the table are chosen from the 12 languages of MKEB. }
    \label{tab:avg-length}
\end{table}
\begin{table}[t]
    \centering
    \scalebox{0.7}{\begin{tabular}[\columnwidth]{c|ccccc}
        \toprule
        \textbf{Datasets} & Edit & Paraphrase & Neighborhood & MQ & languages \\
        \midrule
        CounterFact & $\checkmark$ & $\checkmark$ & $\checkmark$ & & 1 \\
        zsRE & $\checkmark$ & $\checkmark$ & $\checkmark$ & & 1 \\
        MQuAKE & $\checkmark$ & & & $\checkmark$ & 1 \\ 
        MLaKE & $\checkmark$ & & & $\checkmark$ & 5 \\
        \midrule
        MKEB & $\checkmark$ & $\checkmark$ & $\checkmark$ & $\checkmark$ & 12 \\
        \bottomrule
    \end{tabular}}
    \caption{The comparison between our dataset MKEB and other datasets. The number corresponding to Languages represent the count of supported languages.}
    \label{tab:dataset-comparison}
\end{table}
\subsection{Data Statistics}\label{subsec:data-statistics}
Figure~\ref{fig:data-relation} displays the distribution of relationships, while Table~\ref{tab:avg-length} shows the average length of various types of prompts. The MKEB dataset comprises over 9,000 samples, each including different types of prompts in 12 languages. 
Additionally, we present a comparison between MKEB and other datasets, including CounterFact (\citealp{rome}), zsRE (\citealp{zsRE}), MQuAKE (\citealp{MQuAKE}), and MLaKE (\citealp{mlake}), in Table~\ref{tab:dataset-comparison}. 
In summary, our dataset provides a wide range of prompts and multi-hop questions, supporting up to 12 languages. Employing MKEB in research facilitates a comprehensive evaluation of the reliability, generality, locality, cross-lingual transferability of editing algorithms, and multi-hop reasoning capability of edited models.

\section{Methodology}
\begin{figure*}[h!]
    \centering
    \scalebox{0.85}{
        \includegraphics[width=\textwidth]{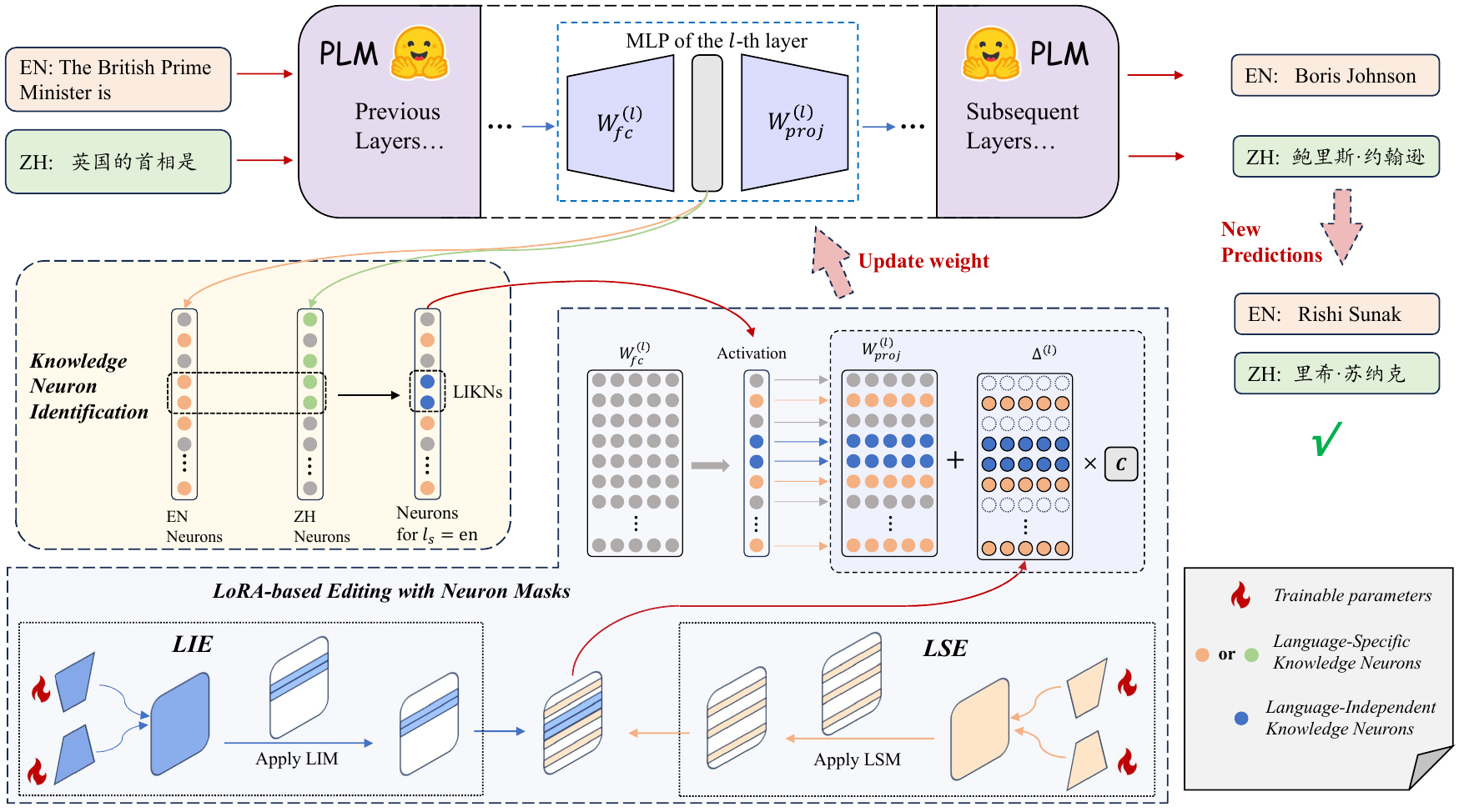}
    }
    \caption{The overall framework of MEMLA, where $W_{fc}^{\left(l\right)}$ and $W_{proj}^{\left(l\right)}$ denote the first and second weights of the MLP in the $l$-th layer, respectively. LIKNs represent language-independent knowledge neurons. LIE and LSE represent the language-independent editor and language-specific editor, respectively. LIM and LSM denote the language-independent neuron mask and the language-specific neuron mask, respectively.}
    \label{fig:method}
\end{figure*}

In this section, we present a detailed introduction to our method, MEMLA, with an overview of the framework depicted in Figure~\ref{fig:method}. Our approach comprises two primary components: 
(1) Knowledge Neuron Identification (\S\ref{subsec:kn-identification}), which leverages integrated gradients (\citealp{integrated-gradients}) to determine the knowledge neurons correlated with a specific fact. 
(2) LoRA-based Editing with Neuron Masks (\S\ref{subsec:mkeditor}), which utilizes editors based on LoRA with neuron masks to selectively modify crucial parameters. 
Each component will be thoroughly elucidated.

\subsection{Knowledge Neuron Identification} \label{subsec:kn-identification}
The probability that the language model predicts the correct answer $y^*$ for an input prompt $x$ can be formally represented as follows:
\begin{equation}
    P_{x}\left(\hat{w}_i^{\left(l\right)} \right)=p\left(y^*|x;w_i^{\left(l\right)}=\hat{w}_i^{\left(l\right)}\right),
\end{equation}
where $w_i^{\left(l\right)}$ denotes the $i$-th intermediate neuron of the MLP in the $l$-th layer, and $\hat{w}_i^{\left(l\right)}$ is the value assigned to $w_i^{\left(l\right)}$. 
The attribution score of each neuron can be computed using integrated gradients:
\begin{equation}
    \footnotesize
    \text{Attr}\left(w_i^{\left(l\right)}\right) = \Delta w_i^{\left(l\right)} \int_{0}^{1}\frac{\partial P_x\left( w_i^{\prime\left(l\right)} + \alpha\cdot \Delta w_i^{\left(l\right)} \right)}{\partial w_i^{\left(l\right)}}\mathrm{d}\alpha,
\end{equation}
where $\Delta w_i^{\left(l\right)}=\Bar{w}_j^{\left(l\right)} - w^{\prime\left(l\right)}_i$, $\Bar{w}_i^{\left(l\right)}$ is the value of $w_i^{\left(l\right)}$, and $w^{\prime\left(l\right)}_i$ is the baseline vector of $w_i^{\left(l\right)}$.
We compute the attribution score and determine the set of knowledge neurons $\mathcal{N}_x^{k}$ for language $k$ following \citet{likn}.
The language-independent knowledge neurons can be derived by intersecting the sets of knowledge neurons across all languages:
\begin{equation}
    \mathcal{I}_{x} = \bigcap_{k=1}^{K}\mathcal{N}_{x}^{k},
    \label{eq:intersection}
\end{equation}
where $\mathcal{I}_{x}$ denotes the language-independent knowledge neurons associated with prompt $x$. Then, we can obtain language-specific knowledge neurons for each individual language:
\begin{equation}
    \mathcal{S}_{x}^{k} = \mathcal{N}_{x}^{k} - \mathcal{I}_{x}.
    \label{eq:dif-set}
\end{equation}

\subsection{LoRA-based Editing with Neuron Masks} \label{subsec:mkeditor}
Existing studies have considered Multi-Layer Perceptrons (MLPs) within LLMs as key-value memories (\citealp{transformer-ff-layers-are-kv}; \citealp{rome}), where the first MLP layer acts as a key, enabling the second MLP layer to generate an appropriate value. 
This paper adheres to this theory and focuses on the parameters of the second MLP layer as the editing target. 
As depicted in Figure~\ref{fig:method}, associated vectors within the MLP's second layer engage with knowledge neurons, thereby supporting the generation of knowledge throughout the forward process.

LoRA (\citealp{loramodule}) is a compute-efficient technique that freezes the model weights and injects trainable rank decomposition matrices into each layer of the model: $\Delta \boldsymbol{W} = \boldsymbol{B}\boldsymbol{A}$, 
where $\boldsymbol{B}\in\mathbb{R}^{d\times r}$, $\boldsymbol{A}\in\mathbb{R}^{r\times k}$, and the rank $r\ll\min\left(d, k\right)$. 
In this paper, we develop two types of editors for modifying a weight $\boldsymbol{W}_{proj}^{\left(l\right)}$:

(1) Language-Specific Editor (LSE), which employs two low-rank matrices to determine the adjustment for monolingual scenarios:
\begin{equation}
    \boldsymbol{\Delta}_s^{\left(l\right)} = \boldsymbol{B}_s^{\left(l\right)}\boldsymbol{A}_s^{\left(l\right)}.
\end{equation}

The ranks of the two matrices can be formulated as: $r_s^{\left(l\right)}=n\cdot \lvert\mathcal{S}_x^k\left(l\right)\rvert$, where $n$ is a positive integer and $\mathcal{S}_x^k\left(l\right)$ is the set of language-specific knowledge neurons in layer $l$.

(2) Language-Independent Editor (LIE), which shares a similar mathematical form with the LSE, i.e.,  $\boldsymbol{\Delta}_i^{\left(l\right)} = \boldsymbol{B}_i^{\left(l\right)}\boldsymbol{A}_i^{\left(l\right)}$.

To accurately update parameters associated with knowledge neurons, we introduce neuron masks that enable LoRA-based editors to selectively update relevant parameters while avoiding changes to unrelated ones. Corresponding to the two editors, we develop Language-Specific neuron Mask (LSM) and Language-Independent neuron Mask (LIM) as follows:
\begin{equation}
\footnotesize
    \begin{aligned}
        \boldsymbol{\Delta}_s^{\left(l\right)} \longleftarrow \left(\boldsymbol{B}_s^{\left(l\right)}\boldsymbol{A}_s^{\left(l\right)}\right)\odot \boldsymbol{M}_s^{\left(l\right)},\\
        \boldsymbol{\Delta}_i^{\left(l\right)} \longleftarrow \left(\boldsymbol{B}_i^{\left(l\right)}\boldsymbol{A}_i^{\left(l\right)}\right)\odot \boldsymbol{M}_i^{\left(l\right)} ,
    \end{aligned}
\end{equation}
where $\odot$ denotes the element-wise product. The mask $\boldsymbol{M}_s^{\left(l\right)}$ is defined as follows:
\begin{equation}
\footnotesize
    \boldsymbol{M}_s^{\left(l\right)}\left[i, :\right] = \left\{
    \begin{array}{rcl}
        \boldsymbol{1}, & & i \in \mathcal{S}_x^k\left(l\right) \\
        \boldsymbol{0}, & & \text{otherwise}
    \end{array}
    \right.,
\end{equation}
and the definition is similar for $\boldsymbol{M}_i^{\left(l\right)}$. Subsequently, we can deduce the final modification:
\begin{equation}
\footnotesize
    \begin{aligned}
        &\boldsymbol{\Delta}^{\left(l\right)} = \boldsymbol{\Delta}_s^{\left(l\right)} + \boldsymbol{\Delta}_i^{\left(l\right)} \\
        &\boldsymbol{W}_{proj}^{\left(l\right)} \longleftarrow \boldsymbol{W}_{proj}^{\left(l\right)} + c\left(\boldsymbol{\Delta}^{\left(l\right)}, \boldsymbol{W}_{proj}^{\left(l\right)}\right)\cdot \boldsymbol{\Delta}^{\left(l\right)},
    \end{aligned}
\end{equation}
where $c$ is a function generating a coefficient that regulates the magnitude of the modification.

\section{Experiments}
\subsection{Experimental Setup}
\paragraph{Metrics.} We evaluate the knowledge editing algorithms in terms of \textbf{reliability}, \textbf{generality}, and \textbf{locality}, which are quantified by Edit Success (ES), Paraphrase Score (PS), and Neighborhood Score (NS), respectively.
Additionally, we compute the \textbf{transferability} of a metric by averaging its values across various source-target language pairs.
A detailed introduction and calculation of these metrics are offered in Appendix~\ref{sec:appendix-metrics}.

\paragraph{Baselines.} We employ the following approaches as baselines:
(1) \textbf{Finetuning} (\textbf{FT}) (\citealp{ft-l}), which fine-tunes the language model with a parameter-space $L_{\infty}$ norm constraint. (2) \textbf{ROME} (\citealp{rome}), which utilizes causal mediation analysis to identify the editing area and updates the weight of the MLP. (3) \textbf{MEMIT} (\citealp{memit}), which builds upon the framework of ROME and enables simultaneous editing for multiple instances. (4) \textbf{MEND} (\citealp{mend}), which employs a hyper network to map a fine-tuning gradient into a new parameter update. (5) \textbf{IKE} (\citealp{iclchangbaobao}), which is based on in-context learning. (6) \textbf{PMET} (\citealp{pmet}), which is based on MEMIT and involves the attention value to achieve a better performance.

\paragraph{Backbone.} Considering that Multilingual Knowledge Editing (MKE) needs to be performed in multiple languages and most existing works are conducted on language models of the GPT series (\citealp{mend}; \citealp{rome}; \citealp{memit}; \citealp{surveyofedit}), we utilize mGPT (\citealp{mgpt}), a multilingual language model with 1.3B parameters, as the backbone for this task.

\begin{table*}[!h]
    \centering
    \scalebox{0.7}{
        \begin{tabular}{c|c|ccccccccccccc}
        \toprule
         \textbf{Metrics} & \textbf{Methods}  & \textbf{en-en} & \textbf{en-zh} & \textbf{en-fr} & \textbf{en-de} & \textbf{en-it} & \textbf{en-es} & \textbf{en-pt} & \textbf{en-pl} & \textbf{en-ru} & \textbf{en-ja} & \textbf{en-ko} & \textbf{en-th} & \textbf{en-avg}\\
         \midrule
           \multirow{7}{*}{\textbf{ES}} & IKE & 99.50 & 94.65 & 93.75 & 97.25 & 95.85 & 94.05 & 95.35 & 96.20 & 94.70 & 93.30 & 93.60 & 67.10 & 92.34 \\
           \cmidrule{2-15}
           & FT & 97.45 & 64.40 & 81.90 & 79.00 & 77.30 & 81.85 & 77.15 & 71.60 & 66.05 & 62.50 & 63.25 & 61.30 & 71.48\\
           & ROME & \textbf{100.00} & 56.20 & 83.20 & 75.45 & 75.05 & 78.00 & 74.95 & 68.45 & 56.00 & 53.80 & 56.55 & 60.00 & 67.06\\
           & MEND & 35.25 & 48.70 & 40.20 & 44.00 & 43.30 & 45.30 & 47.45 & 46.55 & 49.70 & 50.70 & 53.90 & 55.05 & 47.71 \\
           & PMET & 69.60 & 58.30 & 54.70 & 58.65 & 56.60 & 56.3 & 57.35 & 55.30 & 55.75 & 51.85 & 54.10 & 53.15 & 55.64 \\
           & MEMIT & 99.80 & 59.15 & \textbf{88.85} & 79.90 & 78.40 & 81.05 & 77.10 & 75.60 & 59.05 & 56.25 & 59.30 & 60.85 & 70.50 \\
           \cmidrule{2-15}
           & MEMLA (Ours) & 99.85 & \textbf{67.55} & 86.10 & \textbf{82.90} & \textbf{79.90} & \textbf{84.25} & \textbf{78.85} & \textbf{78.60} & \textbf{67.80} & \textbf{64.15} & \textbf{66.25} & \textbf{61.80} & \textbf{74.38} \\
           \midrule
           \multirow{7}{*}{\textbf{PS}} & IKE & 87.64 & 84.05 & 84.57 & 83.85 & 82.97 & 83.22 & 83.78 & 85.95 & 84.35 & 84.37 & 86.27 & 66.99 & 82.76 \\
           \cmidrule{2-15}
           & FT & 86.47 & 63.52 & 70.15 & 68.32 & 67.47 & 71.58 & 69.78 & 63.60 & 62.48 & 60.24 & 61.53 & 59.46 & 65.29 \\
           & ROME & 86.98 & 59.21 & 74.47 & 66.46 & 68.32 & 70.44 & 70.64 & 63.60 & 56.76 & 55.43 & 58.78 & 58.02 & 63.83 \\
           & MEND & 49.97 & 54.33 & 55.01 & 54.06 & 50.98 & 57.19 & 56.37 & 53.51 & 55.29 & 53.68 & 55.86 & 55.24 & 54.68 \\
           & PMET & 64.68 & 56.15 & 54.82 & 59.25 & 56.86 & 56.20 & 58.17 & 55.65 & 54.69 & 51.72 & 53.49 & 52.89 & 55.44 \\
           & MEMIT & 93.35 & 62.59 & 80.87 & 74.22 & 73.63 & 74.89 & 74.02 & 70.02 & 59.38 & 57.26 & 60.71 & 59.63 & 67.93 \\
           \cmidrule{2-15}
           & MEMLA (Ours) & \textbf{95.56} & \textbf{65.89} & \textbf{82.03} & \textbf{79.71} & \textbf{77.03} & \textbf{79.59}& \textbf{76.33} & \textbf{74.52} & \textbf{67.74} & \textbf{61.96} & \textbf{63.62} & \textbf{60.26} & \textbf{71.70} \\
           \midrule
           \multirow{7}{*}{\textbf{NS}} & IKE & 47.27 & 42.93 & 37.33 & 31.18 & 33.05 & 35.01  & 28.20 & 27.35 & 19.54 & 37.57 & 22.16 & 7.02 & 29.21 \\
           \cmidrule{2-15}
           & FT & 15.40 & 11.23 & 8.46 & 8.65 & 5.88 & \textbf{9.96} & 6.29 & 7.84 & 4.69 & 6.29 & 13.58 & 8.23 & 8.28 \\
           & ROME & 16.14 & 11.38 & 7.68 & 9.83 & 5.43 & 8.84 & 5.93 & 7.66 & 4.92 & \textbf{7.92} & 13.61 & 8.34 & 8.32 \\
           & MEND & 11.53 & 11.33 & 5.38 & 6.28 & 4.33 & 6.68 & 4.96 & 6.24 & 4.51 & 7.22 & 12.69 & 7.21 & 6.98 \\
           & PMET & 5.99 & 0.00 & 0.00 & 0.00 & 0.00 & 0.00 & 0.00 & 0.00 & 0.00 & 0.00 & 0.00 & 0.00 & 0.00 \\
           & MEMIT & 17.17 & 11.55 & 8.06 & 10.33 & 5.66 & 9.08 & 6.18 & 7.56 & 5.05 & 7.83 & 14.05 & \textbf{8.35} & 8.52 \\
           \cmidrule{2-15}
           & MEMLA (Ours) & \textbf{17.91} & \textbf{11.68} & \textbf{10.26} & \textbf{11.13} & \textbf{6.71} & 9.41 & \textbf{6.34} & \textbf{9.42} & \textbf{5.28} & 7.14 & \textbf{14.36} & 7.28 & \textbf{9.00} \\
           \bottomrule
        \end{tabular}
    }
    \caption{Corresponding results in the case of editing with English. en-zh indicates that the source language is English and the target language is Chinese; the same applies to the rest. en-avg represents the average performance for cross-lingual scenarios (i.e., transferability).}
    \label{tab:en-edit}
\end{table*}

\begin{table*}[!h]
    \centering
    \scalebox{0.7}{
        \begin{tabular}{c|c|ccccccccccccc}
        \toprule
         \textbf{Metrics} & \textbf{Methods}  & \textbf{zh-en} & \textbf{zh-zh} & \textbf{zh-fr} & \textbf{zh-de} & \textbf{zh-it} & \textbf{zh-es} & \textbf{zh-pt} & \textbf{zh-pl} & \textbf{zh-ru} & \textbf{zh-ja} & \textbf{zh-ko} & \textbf{zh-th} & \textbf{zh-avg}\\
         \midrule
           \multirow{7}{*}{\textbf{ES}} & IKE & 93.85 & 96.10 & 88.30 & 92.45 & 91.30 & 90.25 & 90.60 & 92.50 & 87.30 & 87.10 & 86.50 & 65.05 & 87.75 \\
           \cmidrule{2-15}
           & FT & 53.05 & 95.25 & 54.20 & 56.10 & 55.60 & 58.90 & 60.25 & 57.15 & 58.15 & 60.50 & 62.85 & 60.80 & 57.96 \\
           & ROME & 52.95 & 98.55 & 56.55 & 55.90 & 56.05 & 58.00 & 59.00 & 57.25 & 53.65 & 59.20 & 61.30 & 59.65 & 57.23 \\
           & MEND & 42.05 & 37.55 & 44.55 & 47.05 & 46.35 & 48.20 & 48.65 & 44.30 & 49.10 & 52.20 & 51.25 & 53.45 & 47.92 \\
           & PMET & 56.70 & 57.95 & 53.85 & 53.95 & 52.90 & 56.35 & 57.75 & 55.05 & 57.80 & 58.15 & 54.75 & 53.45 & 55.52 \\
           & MEMIT & 55.75 & 99.45 & 57.70 & 57.20 & 55.80 & 58.60 & 59.90 & 58.05 & 54.50 & 62.55 & 63.25 & 59.80 & 58.46 \\
           \cmidrule{2-15}
           & MEMLA (Ours) & \textbf{64.60} & \textbf{100.00} & \textbf{64.35} & \textbf{65.15} & \textbf{64.65} & \textbf{65.25} & \textbf{66.00}  & \textbf{63.20} & \textbf{68.45} & \textbf{67.90} & \textbf{69.55} & \textbf{61.60}& \textbf{65.52} \\
           \midrule
           \multirow{7}{*}{\textbf{PS}} & IKE & 86.32 & 87.48 & 82.08 & 82.85 & 81.05 & 82.00 & 82.79 & 82.15 & 78.98 & 79.71 & 80.09 & 64.93 & 80.27 \\
           \cmidrule{2-15}
           & FT & 57.75 & 87.38 & 57.66 & 57.46 & 54.53 & 59.65 & 61.08 & 55.36 & 58.46 & 59.45 & 60.58 & 58.54 & 58.23 \\
           & ROME & 56.22 & 95.59 & 59.08 & 57.65 & 56.52 & 59.46 & 61.82 & 56.42 & 55.44 & 60.34 & 62.83 & 58.10 & 58.53 \\
           & MEND & 45.99 & 48.75 & 52.58 & 50.97 & 49.07 & 55.41 & 57.36 & 50.55 & 48.83 & 51.89 & 52.78 & 54.92 & 51.85 \\
           & PMET & 57.63 & 56.89 & 59.89 & 56.13 & 52.58 & 57.21 & 58.20 & 55.51 & 57.55 & 58.05 & 53.42 & 50.19 & 56.03 \\
           & MEMIT & 56.96 & 96.94 & 59.43 & 58.00 & 57.41 & 59.23 & 61.82 & 56.84 & 55.49 & 63.49 & 65.15 & 58.84 & 59.33 \\
           \cmidrule{2-15}
           & MEMLA (Ours) & \textbf{69.44} & \textbf{99.80} & \textbf{65.77} & \textbf{68.01} & \textbf{65.59} & \textbf{68.52} & \textbf{68.01} & \textbf{63.32} & \textbf{66.75} & \textbf{66.99} & \textbf{68.89} & \textbf{59.92} & \textbf{66.47} \\
           \midrule
           \multirow{7}{*}{\textbf{NS}} & IKE & 40.75 & 39.73 & 31.70 & 24.54 & 25.90 & 28.80 & 21.93 & 19.93 & 13.79 & 31.69 & 14.65 & 7.42 & 23.74 \\
           \cmidrule{2-15}
           & FT & 16.36 & 20.84 & \textbf{6.88} & \textbf{10.12} & 4.75 & 8.68 & 5.58 & 7.22 & \textbf{4.70} & 22.75 & 13.37 & 9.02 & 9.95 \\
           & ROME & 16.33 & 17.51 & 6.53 & 9.61 & 4.43 & 8.30 & 5.43 & 7.01 & 4.43 & 10.22 & 13.32 & 8.22 & 8.53 \\
           & MEND & \textbf{19.21} & 8.21 & 6.53 & 9.69 & \textbf{5.04} & \textbf{9.71} & \textbf{6.45} & \textbf{9.18} & 4.34 & 6.93 & 11.32 & \textbf{9.13} & 8.87 \\
           & PMET & 1.07 & 7.63 & 0.00 & 0.01 & 0.02 & 0.00 & 0.01 & 0.01 & 0.00 & 28.80 & 0.00 & 2.92 & 2.98 \\
           & MEMIT & 16.87 & 21.41 & 6.77 & 10.00 & 4.71 & 8.41 & 5.50 & 7.40 & \textbf{4.70} & 13.94 & \textbf{13.75} & 8.39 & 9.13 \\
           \cmidrule{2-15}
           & MEMLA (Ours) & 14.77 & \textbf{43.57} & 6.54 & 8.98 & 4.81 & 7.36 & 5.33 & 5.43 & 3.49 & \textbf{43.20} & 3.88 & 7.10 & \textbf{10.08} \\
           \bottomrule
        \end{tabular}
    }
    \caption{Corresponding results in the case of editing with Chinese. zh-en indicates that the source language is English and the target language is Chinese; the same applies to the rest. zh-avg represents the average performance for cross-lingual scenarios (i.e., transferability).}
    \label{tab:zh-edit}
\end{table*}

\subsection{Main Results}
The main results when the source language is English and Chinese are illustrated in Table~\ref{tab:en-edit} and Table~\ref{tab:zh-edit}, respectively.
These results yield several significant observations:

(1) Our approach outperforms other methods that modify model parameters. Specifically, when the source language is Chinese, the ES and PS for transferability (zh-avg) exhibit improvements of 7.06\% and 7.14\%, respectively, compared to MEMIT. This highlights the efficacy of the proposed MEMLA for this task.

(2) MEMLA exhibits a notable capacity to preserve irrelevant knowledge. This is demonstrated when examining the zh-zh and zh-ja settings, where the NS of MEMLA increased by 22.16\% and 29.26\%, respectively, compared to MEMIT. We attribute this to MEMLA modifying only parameters associated with knowledge neurons, thereby facilitating the preservation of unrelated knowledge.

(3) IKE achieves exceptional performance through in-context learning. 
We contend that this method merely induces the model to generate new responses by appending a new fact to the prompt without actually altering the model's internal knowledge.
Therefore, we regard it as a theoretical upper bound for editing performance.

\subsection{Ablation Study}
We conduct an ablation study as follows: (1) without the language-specific neuron mask (w/o LSM), which eliminates the LSM and employs the LoRA module directly for weight updates; (2) without the language-specific editor (w/o LSE), which removes the LSE and relies entirely on the LIE for editing; (3) without the language-independent neuron mask (w/o LIM), which deactivates the LIM and uses the LoRA module directly for altering weights; (4) without the language-independent editor (w/o LIE), which disables the LIE and relies exclusively on the LSE for knowledge editing. 
\begin{table}[H]
\centering
\scalebox{0.7}{\begin{tabular}{l|ccc|ccc}
\toprule
\multirow{2}{*}{\textbf{Methods}} & \multicolumn{3}{c|}{\textbf{zh-zh}} & \multicolumn{3}{c}{\textbf{zh-avg}} \\
\cmidrule{2-4} \cmidrule{5-7}
 & ES & PS & NS  & ES & PS & NS \\
\midrule
\textbf{MEMLA} & \textbf{100} & \textbf{99.80}  & \textbf{43.57}  & \textbf{65.52} & \textbf{66.47} & \textbf{10.08}  \\
\midrule
w/o LSM & 97.6 & 97.34 & 30.38  & 62.60 & 62.91 & 2.57  \\
w/o LSE  & 90.11 & 86.27 & 37.08 & 60.86 & 63.05 & 9.48 \\
w/o LIM & 93.2 & 86.91 & 31.59  & 62.90 & 64.17 & 4.15  \\
w/o LIE & 95.39 & 93.53 & 30.24 & 61.20 & 62.46 & 9.98 \\
\bottomrule
\end{tabular}}
\caption{
The ablation results of our approach.
}
\label{tab:ablation}
\end{table}
The results are presented in Table~\ref{tab:ablation}, from which we can draw the following key conclusions:

(1) Effectiveness of neuron masks. The elimination of either the LIM or the LSM results in a decline in the model's performance in both monolingual and cross-lingual scenarios. This indicates that the neuron masks improve editing precision and enable localized parameter adjustments, thereby enhancing performance.

(2) Effectiveness of LSE and LIE. When either is eliminated, performance significantly declines, implying that the two types of editors for the task are highly effective. 

\begin{table*}[h]
    \centering
    \scalebox{0.9}{
    \begin{tabular}{ccccccc}
    \toprule
       \textbf{Method}  & \textbf{en-en} & \textbf{en-avg} & \textbf{zh-zh} & \textbf{zh-avg} & \textbf{fr-fr} & \textbf{fr-avg} \\
       \midrule
        FT & 1.82 & 1.29 & 9.43 & 2.00 & 2.33 & 1.15 \\
        ROME & 0.88 & 1.28 & 7.61 & 1.42 & 1.33 & 1.16 \\
        MEMIT & 2.22 & 1.59 & 16.50 & 1.74 & 3.38 & 1.27  \\
        \midrule
        \multirow{2}{*}{MEMLA} & \textbf{13.72} & \textbf{3.35} & \textbf{82.69} & \textbf{5.62} & \textbf{27.28} & \textbf{2.50} \\
        & ($\uparrow$ 518.02\%) & ($\uparrow$ 110.69\%) & ($\uparrow$ 401.15\%) & ($\uparrow$ 222.99\%) & ($\uparrow$ 707.10\%) & ($\uparrow$ 96.85\%) \\
        \bottomrule
    \end{tabular}
    }
    \caption{Performance of the edited model answering multi-hop questions. ($\uparrow$) represents the improvements over the previous state-of-the-art method MEMIT.}
    \label{tab:mq}
\end{table*}

\begin{figure*}[h]
    \centering
    \scalebox{0.9}{\subfigure[Performance on ANLI.] {
        \includegraphics[width=0.48\textwidth]{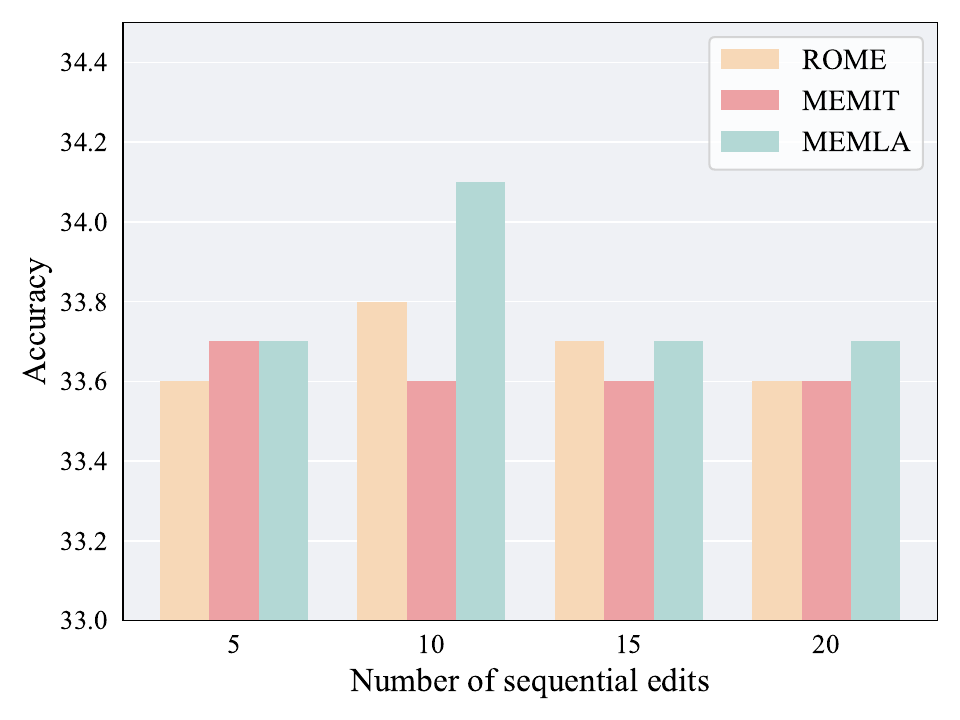}
    }
    \hspace{10mm}
    \subfigure[Performance on PIQA.]{
        \includegraphics[width=0.48\textwidth]{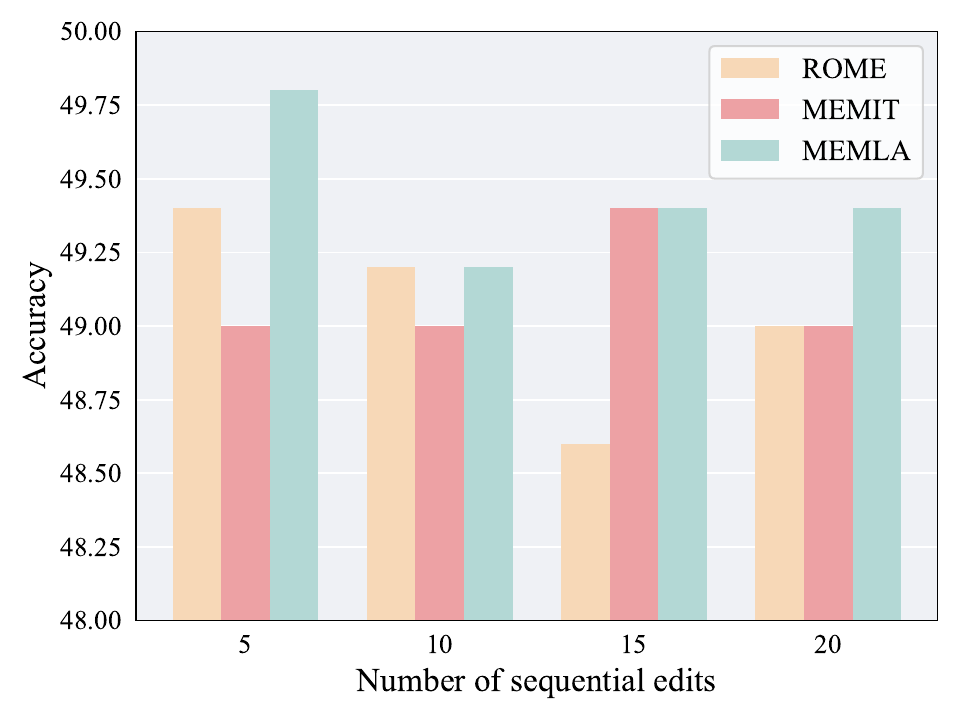}
    }}
    \caption{Performance of the edited model on downstream tasks.}
    \label{fig:task}
\end{figure*}

\subsection{Multi-hop Reasoning Capability of the Edited Model}
Further research is needed to determine if the model has effectively integrated the revised knowledge and fully comprehended the additional knowledge implied by the initial edit. 
This evaluation component is alternatively referred to as \textit{ripple effects} (\citealp{ripple-effects}).
In this study, we employ multi-hop questions to evaluate the model's capability to acquire implicit knowledge via multi-hop reasoning. 
The definition and computation of the evaluation metric are detailed in Appendix~\ref{sec:appendix-metrics}, and the results are shown in Table~\ref{tab:mq}. 
From the results, we can clearly see that our approach demonstrates superior performance compared to other methods in all settings. Notably, in monolingual settings such as en-en and fr-fr, the multi-hop reasoning capability of the model edited by MEMLA has been substantially enhanced, resulting in improvements over MEMIT of 518.02\% and 707.10\%, respectively. Furthermore, MEMLA proves effective in cross-lingual situations, with improvements of 110.69\% and 222.99\% in en-avg and zh-avg, respectively.

We believe that MEMLA enhances the multi-hop reasoning capability of the edited model by performing editing at a more general level, updating crucial parameters associated with knowledge neurons, and thus facilitating the integration of new knowledge.

\subsection{Impact of Knowledge Editing on Language Model}
Knowledge editing inevitably impacts the general capabilities of the model (\citealp{butterfly}; \citealp{unveiling}; \citealp{editing-can-hurt}; \citealp{missing-piece}). 
To explore these impacts, we apply the edited model to downstream tasks such as ANLI (\citealp{anli}) and PIQA (\citealp{piqa}) and assess its performance on these tasks. 
The results are presented in Figure~\ref{fig:task}.

The model's performance, after being edited using MEMLA, evidently surpasses that of ROME and MEMIT on downstream tasks. 
We attribute this improvement to the application of LoRA-based editors with neuron masks, which allows for more precise, flexible, and lightweight editing, thereby reducing potential damage to the model. 

\section{Conclusion}
In this paper, we introduce the multilingual knowledge editing benchmark (MKEB), which covers 12 languages and provides a comprehensive evaluation framework for reliability, generality, locality, transferability of editing algorithms, and multi-hop reasoning capability of edited models. 
Furthermore, we propose an effective multilingual knowledge editing method based on LoRA with neuron masks (MEMLA). To improve editing precision, we identify two categories of knowledge neurons. To efficiently update parameters and facilitate the propagation of updates across multiple languages, we create neuron masks for LoRA to adjust critical parameters. 
Experimental results indicate that our approach outperforms other baselines, enabling the edited model to capture additional implicit knowledge through multi-hop reasoning while minimally impacting the model's general performance on downstream tasks.

\section*{Limitations}
Due to computational resource constraints, we have not yet explored multilingual knowledge editing on larger models. 
Moreover, some LLMs, such as GPT-4 (\citealp{gpt4}), are ``black box'' models with undetectable internal structures. Consequently, the mechanism of knowledge sharing between diverse languages within these models remains unclear. Further discussion is required to determine how to perform multilingual knowledge editing on these models in future work.

\bibliography{acl_latex}

\appendix
\section{Dataset}
\label{sec:appendix-dataset}

We have developed the following guidelines to induce ChatGPT to generate various prompts and multi-hop questions:

(1) To accommodate algorithms like ROME and MEMIT, ``\verb|{}|'' is utilized within the edit prompt to serve as a placeholder for the subject.

(2) A paraphrase prompt is designed to restate the edit prompt while preserving its original meaning.
    
(3) The subject of the generated neighborhood prompt should be similar yet distinct. For instance, if the edit prompt is ``The capital of China is'', the subject of the neighborhood prompt should ideally fall within the category of countries (e.g., Japan, Australia, etc.).

(4) For a knowledge chain consisting of multiple triples, the generated multi-hop question should exclude the intermediate entities. For example, given a knowledge chain (Ubisoft, country, France), (France, capital, Paris), the appropriate multi-hop question would be formulated as ``What is the capital of the country to which Ubisoft belongs?''. The question avoids mentioning the intermediate entity (France) and queries about the head entity of the first triple, with the answer being the tail entity of the last triple.

These guidelines are incorporated into the input prompts for ChatGPT, enabling it to generate instances consistent with our expectations.

\section{Metrics}
\label{sec:appendix-metrics}

In multilingual settings, we typically use $l_s$ as the source language for editing and $l_t$ as the target language for assessment.

\textbf{Reliability} measures whether the new knowledge edited by $l_s$ has been integrated into the knowledge set of $l_t$ within the model through the edit prompt. It is quantified by Edit Success (\textbf{ES}) in $l_t$ and is calculated as follows:
\begin{equation}
\footnotesize
    \mathrm{ES}_{l_s,l_t} = \mathbb{E}_{x\in S_e\left(l_t\right)}\left[ \mathds{1}\left(P_{l_s}\left(y^*|x\right)>P_{l_s}\left(y^o|x\right)\right)\right],
\end{equation}
where $x$ represents the edit prompt, $S_e\left(l_t\right)$ denotes the collection of all edit prompts corresponding to $l_t$, $P_{l_s}$ represents the output probability of the model edited by $l_s$, $y^*$ denotes the new answer for $x$, and $y^o$ denotes the original answer. $\mathds{1}\left(\cdot\right)$ is the indicator function.
Evidently, when $l_s=l_t$, this metric can assess the success rate of monolingual editing.

\textbf{Generality} refers to the ability of the edited model to generate the desired output consistently across various prompts that convey the same meaning (i.e., paraphrases). Generality can be quantified by the Paraphrase Score (\textbf{PS}):
\begin{equation}
\footnotesize
    \mathrm{PS}_{l_s,l_t} = \mathbb{E}_{x\in S_p\left(l_t\right)}\left[ \mathds{1}\left(P_{l_s}\left(y^*|x\right)>P_{l_s}\left(y_0|x\right)\right)\right],
\end{equation}
where $x$ denotes the paraphrase prompt in language $l_t$ and $S_p\left(l_t\right)$ represents the collection of all paraphrase prompts in language $l_t$.

\textbf{Locality} reflects the ability of the edited model to retain its original irrelevant knowledge, as measured by the Neighborhood Score (\textbf{NS}). We treat the prediction $f_{l_e}\left(x\right)$ of the edited model and the ground truth $y^*$ as bags of tokens and compute the average F1 score for them as NS:
\begin{equation} \label{eq:ns}
     \mathrm{NS}_{l_s,l_t} = \mathbb{E}_{x\in S_n\left(l_t\right)} \mathrm{F1}\left( f_{l_s}\left(x\right),\ y^* \right),
\end{equation}
where $x$ denotes the neighborhood prompt, $S_n\left(l_t\right)$ represents the collection of all neighborhood prompts in language $l_t$, $f_{l_s}\left(x\right)$ denotes the prediction of the language model edited by $l_s$, and $y^*$ denotes the ground truth for the input prompt $x$.

\textbf{Transferability} refers to the cross-linguistic adaptation of the three aforementioned metrics. The transferability of a metric is calculated by averaging its values across various source-target language pairs. 
In Table~\ref{tab:en-edit} and Table~\ref{tab:zh-edit}, the cross-lingual transferability is listed as en-avg and zh-avg, respectively. These terms represent the average performance of the aforementioned metrics for several language pairs where $l_s$ and $l_t$ are not identical.

In this paper, we evaluate the multi-hop reasoning capability of the edited model using multi-hop questions. The corresponding evaluation metric is the Question Score (\textbf{QS}), which is calculated similarly to the NS (Equation \ref{eq:ns}):
\begin{equation}
    \mathrm{QS}_{l_s,l_t} = \mathbb{E}_{x\in S_q\left(l_t\right)} \mathrm{F1}\left( f_{l_s}\left(x\right),\ y^* \right),
\end{equation}
where $S_q\left(l_t\right)$ denotes the set of multi-hop questions, $f_{l_s}\left(x\right)$ denotes the prediction of the language model edited by $l_s$, and $y^*$ denotes the ground truth for the input question $x$.

\end{document}